\documentclass[conference]{IEEEtran}

\IEEEoverridecommandlockouts
\usepackage{cite}
\usepackage{amsmath,amssymb,amsfonts}
\usepackage{algorithmic}
\usepackage{graphicx}
\usepackage{textcomp}
\usepackage{xcolor}
\usepackage{booktabs}

\def\BibTeX{{\rm B\kern-.05em{\sc i\kern-.025em b}\kern-.08em
    T\kern-.1667em\lower.7ex\hbox{E}\kern-.125emX}}

\usepackage{subfigure}

\ifCLASSINFOpdf
\else
\fi
\begin{document}

\title{Ranking Micro-Influencers: a Novel Multi-Task Learning and Interpretable Framework}




\author{\IEEEauthorblockN{Adam Elwood}
\IEEEauthorblockA{\textit{lastminute.com} \\
Chiasso, Switzerland \\
adam.elwood@lastminute.com}
\and
\IEEEauthorblockN{Alberto Gasparin}
\IEEEauthorblockA{\textit{lastminute.com} \\
Chiasso, Switzerland \\
alberto.gasparin@lastminute.com}
\and
\IEEEauthorblockN{Alessandro Rozza}
\IEEEauthorblockA{\textit{lastminute.com} \\
Chiasso, Switzerland \\
alessandro.rozza@lastminute.com}
}

\newcommand{\real}{{\rm I\!R}}
\newcommand{\dv}{d_v}
\newcommand{\dt}{d_t}
\newcommand{\dvt}{(d_v+d_t)}

\maketitle

\begin{abstract}
With the rise in use of social media to promote branded products, the demand for effective influencer marketing has increased. Brands are looking for improved ways to identify valuable influencers among a vast catalogue; this is even more challenging with “micro-influencers”, which are more affordable than mainstream ones but difficult to discover. In this paper, we propose a novel multi-task learning framework to improve the state of the art in micro-influencer ranking based on multimedia content. Moreover, since the visual congruence between a brand and influencer has been shown to be good measure of compatibility, we provide an effective visual method for interpreting our models’ decisions, which can also be used to inform brands' media strategies. We compare with the current state-of-the-art on a recently constructed public dataset and we show significant improvement both in terms of accuracy and model complexity. The techniques for ranking and interpretation presented in this work can be generalised to arbitrary multimedia ranking tasks that have datasets with a similar structure.
\end{abstract}



\section{Introduction}

As people spend increasingly more time on the internet, brands are keen to make use of online marketing platforms \cite{zietek2016influencer}. When connected to search and social media, these platforms are able to personalise adverts, increasing the probability that consumers will interact with them~\cite{inf_zm7}. This has been effective enough for many of the websites on the internet to be mainly funded through advertising revenues. However, as time has progressed, consumers have become desensitised to online advertisements, increasing their use of ad-blocking software and their suspicion of online adverts \cite{15MIR,inf_review,bad_marketing}. 

To find a solution to this issue, brands have started forming partnerships with ``influencers'', social media accounts that have a significant and dedicated following~\cite{inf_zm1,inf_zm2}. Unlike celebrity sponsorship, influencer marketing can have the advantage of costing less and being able to target very specific demographics~\cite{inf_review}. Influencers also typically have an intimate relationship with their audiences, making them a more trusted source of information and more effective advertisers. With the continuous rise in the use of social media platforms, this new form of marketing has had significant success~\cite{abidin2015communicative}.

Despite the promise of this new style of marketing, finding appropriate influencers to advertise their product is a major challenge for brands that want to engage in it \cite{woods2016sponsored}. In contrast to the relatively limited pool of online marketing platforms and celebrities, there are many more individuals that can be defined as influencers~\cite{inf_v_celeb}. Particularly enticing are ``micro-influencers'', those with between 5000 and 100~000 social media followers~\cite{inf_micro}. Due to their relative obscurity, their advertising services can be obtained cheaply, while being very effective in the case that their audience is a good match for a brand. At this point there are hundreds of thousands of micro-influencers, so the challenge is sorting through them and finding the most relevant candidates for a particular brand.

This problem has been made tractable with recent advances in machine learning for computer vision and natural language processing, along with the large quantity of data produced by the social media platforms. One promising technique for ranking micro-influencers based on such techniques was introduced by Gan~\emph{et~al.}~\cite{MIR}. Influencers are ranked for brands, based on the similarity of the images and text in influencer and brand posts, which are pooled and embedded via a custom neural network. This methodology is backed up with further research into influencer marketing, which has helped to support the hypothesis that the visual congruence between a brand and influencer is a good measure of compatibility \cite{inf_dl1}. Other relevant literature is covered in Section~\ref{sec:related}.

In this work we present significant improvements on the algorithm presented by Gan~\emph{et~al.} \cite{MIR} through the introduction of novel neural network architectures, described in Section~\ref{sec:method}. 

To develop these methods, we make use of a dataset of brands and influencers taken from Instagram, originally introduced in \cite{MIR} and described further in Section~\ref{sec:dataset}. The details of the experiments run to demonstrate the efficacy of our methods can be found in Section~\ref{sec:full_exp}.

We also introduce novel methods for the interpretability of influencer ranking models in Section~\ref{sec:interpretation}. This allows the techniques introduced to provide brands with tools for better understanding the impact of their media content, along with the potential for ranking micro-influencers at scale. Recent research in the interpretability of neural networks can be used to understand the exact components of influencer visual content that make them suitable for a certain brand. This allows brands to better tailor their media content to suit specific audiences. 

Finally, we summarise our findings and suggest avenues for future work in Section~\ref{sec:conclusions}.

The main contributions of this work are as follows:
\begin{itemize}
    \item Significant improvements over the state-of-the-art of micro-influencer ranking across major metrics through the addition of multi-task learning to the technique presented in \cite{MIR}.
    \item Novel neural network architectures for ranking, based around a trainable inner product, improving performance with a significant reduction in the number of model parameters.
    \item The introduction of techniques for interpreting the ranking results to provide brands with tools for improving their media content. 
\end{itemize}

\section{Related work}
\label{sec:related}
\subsection{Influencer marketing}

As influencer marketing has grown in popularity, there has been significant interest from the traditional marketing research community \cite{ zykun2020branding,inf_zm3,inf_zm4,inf_zm5,inf_zm6}. This has led to it becoming an established part of the marketing strategy of many big brands. Due to the challenges of finding appropriate influencers detailed above, there has also been an increase in interest in using deep learning algorithms to identify influencers and analyse their interactions with their audience \cite{inf_dl1}. 

The most directly relevant work on influencer ranking with deep learning, and the inspiration for this work, comes from Gan~\emph{et~al.}~\cite{MIR}, which includes a summary of the relevant learning to rank techniques that have been applied here \cite{liu2011learning, li2011short, listnet, luo2015stochastic}. Additionally, Aleksandr~\emph{et~al.}~\cite{farseev2018somin} presented a technical demonstration of an influencer discovery marketplace, SoMin, although did not provide any concrete methods. On top of this, Gelli~\emph{et~al.} proposed a framework to predict the popularity of social images~\cite{gelli2015image}, along with a learn to rank technique designed to help brands find relevant media content~\cite{gelli2018beyond}.

\subsection{Multi-task feature learning}

In multi-task learning, deep neural networks are trained to be able to perform well simultaneously in multiple different, yet related, domains~\cite{caruanamt, zhang2017survey,caruana1997multitask}. Provided there is a latent feature space that is relevant to all the tasks, multi-task learning can help networks to learn more robust features during the optimisation process. This often leads to improved performance in the main task, despite good performance in auxiliary tasks not being strictly necessary. 

There are several different approaches to multi-task learning~\cite{zhang2017survey}, but the most relevant to our use case is high-level feature learning. In early versions of this approach, the learning of different tasks is decoupled by learning the feature covariance for all the tasks, obtaining task specific hidden representations within a shallow network~\cite{evgeniou2007multi,argyriou2008convex}. In the deep learning setting, a common approach is for the different tasks to share several of the first hidden network layers, with only one or two dense layers used for task-specific parameters \cite{zhang2014facial, mrkvsic2015multi, li2014heterogeneous, misra2016cross}. Due to its general purpose applicability, multi-task learning can be useful when training many different deep learning architectures, from image processing convolutional networks~\cite{tellez2020extending} to graph neural networks~\cite{manessi2020graph}.  

\subsection{Deep neural network interpretability}

Deep neural networks are very powerful when applied to computer vision and natural language applications, but act as black-boxes by their nature. This makes it difficult to interpret why they make decisions relevant to the task they are trained to perform on. This has led to increased interest in finding ways to interpret why the networks make the decisions they do~\cite{fan2020interpretability,MONTAVON20181}. Many of the techniques for interpretation are best applied to individual examples, for example providing insight into which parts of an image were most relevant to a convolutional neural network deciding on a certain classification. One of the most useful techniques for computer vision applications is Grad-CAM~\cite{selvaraju2017grad}, which uses the gradients of a target neural network component flowing into the final convolutional layer to highlight the most important regions of the original image. This kind of interpretation is valuable in the influencer marketing use case, as it allows brands to understand the visual content that led to them being matched to certain influencers. Along with checking that appropriate visual content is being used, this information can also be valuable to help the brand design future marketing strategies.

\section{Method}
\label{sec:method}

\subsection{Problem formulation}
Given a set of $n$ brands $\mathcal{B} = \{\mathbf{b}_1, \mathbf{b}_2, \dots, \mathbf{b}_n\}$, with $\mathbf{b}_{j} \in \real^{(\dv+\dt)}$ and a set of $m$ influencers $\mathcal{I} = \{\mathbf{i}_1, \mathbf{i}_2, \dots, \mathbf{i}_m\}$ with $\mathbf{i}_{k} \in \real^{(\dv+\dt)}$, both coming with visual ($d_v$) and textual features ($d_t$), our goal is to recommend a given brand $\mathbf{b}$ with a list of candidate micro-influencers.
We assume that for each brand $\mathbf{b}_*$ a list of associated influencers is available. In the following, we'll refer to these as positive examples for the given brand and describe the set as $\mathcal{I}_+(\mathbf{b}_*)$. In similar fashion, we'll refer to all non-associated micro-influencers, for a given brand $\mathbf{b}_*$, as negative examples, $\mathcal{I}_-(\mathbf{b}_*)$.
Each brand and micro-influencer comes with both textual and visual information for each of its posts. We represent a brand $\mathbf{b} = [\mathbf{b}_{t},\mathbf{b}_{v}]$, where $\mathbf{b}_{t} \in \real^{\dt}$ is a pooled summary of the textual features, obtained by a neural network embedding of word tokens via the spaCy library (based on Tok2Vec) \cite{spacy}, while $\mathbf{b}_{v} \in \real^{\dv}$ is a pooled summary of the visual features obtained from a pretrained VGG-16 \cite{vgg} used as a feature extractor. 
More formally, given the last $N_p$ posts by a given brand, a pooling operation $p: \real^{N_p \times d} \mapsto \real^{d}$ is applied on the image (text) embeddings in order to obtain a unique account representation.
In this work, the pooling operation is a simple average over the posts dimension.
The preprocessing steps described above holds both for brand and micro-influencer accounts, and are deeply inspired by \cite{MIR}.

Given a brand and a set of micro-influencers ($\mathbf{b}$,$\{\mathbf{i}_1,\dots, \mathbf{i}_k\}$) the objective is to provide a ranking (induced by a scoring function) of these micro-influencer w.r.t. the given brand.
In the remaining sections we will describe two approaches to assign a score to each influencer, given a brand, and induce the previously described ranking.

\subsection{Bilinear Similarity}
\label{sec:wsim}
\begin{figure}[ht]
    \centering
    \includegraphics[width=\linewidth]{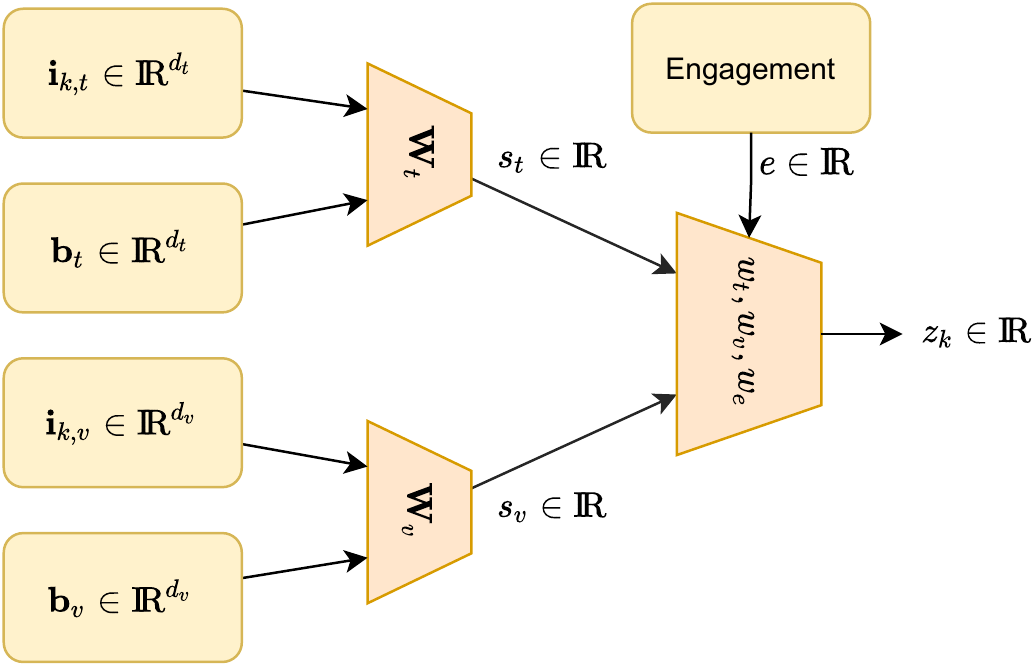}
    \caption{WSim architecture with its parameters $\mathbf{\Theta}=[\mathbf{W}_t, \mathbf{W}_v, w_t,w_v,w_e]$. The input to the model is the pooled summary of the textual and visual features (obtained via VGG-16 and spaCy) for both brand and influencers.}
    \label{fig:wsim}
\end{figure}
The first model we propose is a straightforward generalization of the cosine similarity model. In particular, given a pair of brand and micro-influencer ($\mathbf{b} \in \real^{\dvt}, \mathbf{i}_k \in \real^{\dvt}$) we compute two similarity scores for the visual and textual features separately:
\begin{align}
    s_t &= \mathbf{b}_t^T\mathbf{W}_t\mathbf{i}_{k,t} \\
    s_v &= \mathbf{b}_v^T\mathbf{W}_v\mathbf{i}_{k,v}    
\end{align}
with $\mathbf{W}_t \in \real^{\dt\times \dt}, \mathbf{W}_v \in \real^{\dv \times \dv}$ being diagonal matrices whose weights are the free-parameters of the model. It is worth noting that this bilinear similarity model generalizes the dot product, allowing one to compute a similarity score which is based on the minimization (maximization) of a principled objective. 
The text and visual scores are then combined with a precomputed engagement score $e$ that takes into account the popularity of the posts of a given brand (micro-influencer) as in \cite{MIR}.
A convex combination of the three scores is considered to compute the final score $z_k$ of a micro-influencer $\mathbf{i}_k$ given a brand $\mathbf{b}$:
\begin{equation}
    z_k = w_ts_t + w_vs_v + w_ee \quad \text{s.t. } w_t+w_v+w_e = 1
    \label{eq:z}
\end{equation}
The model's parameter $\mathbf{\Theta}=[\mathbf{W}_t, \mathbf{W}_v, w_t,w_v,w_e]$ are learned by minimizing the ranking objective described in Section \ref{sec:training}; for further details on the optimization strategy we refer the reader to Section \ref{sec:exp}.
In the remaining of the paper we will refer to this model as \textbf{WSim} and its architecture is depicted in Figure \ref{fig:wsim}.

\subsection{Multi-task Learning}
\begin{figure*}[ht]
    \centering
    \includegraphics[width=\linewidth]{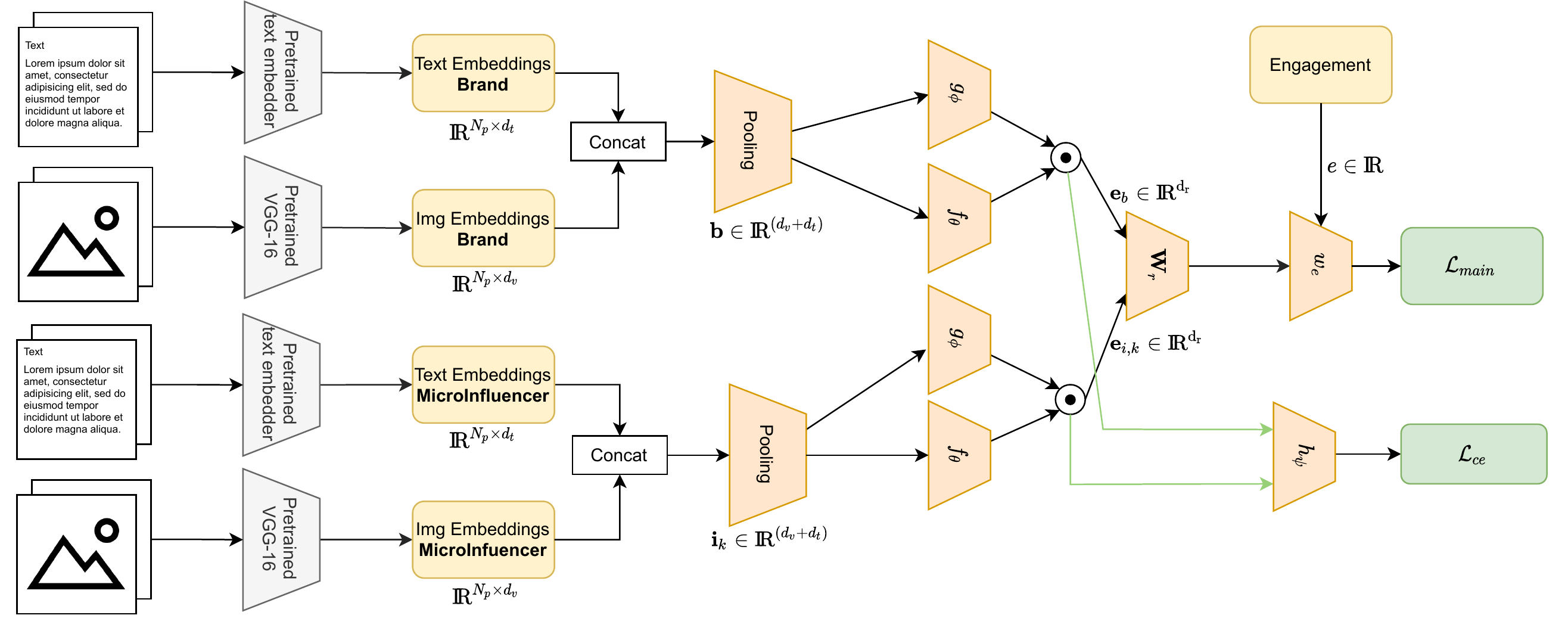}
    \caption{Overall architecture for the multi-task model \textbf{WSim-MT}.}
    \label{fig:mt}
\end{figure*}

The previously presented model builds on top of the feature representations of general-purpose pretrained models such as VGG-16 and the language model provided by spaCy.
In order to further exploit the information available within the dataset and to enhance the learned representation for both influencers and brands, we first feed the pooled text embeddings to a neural network $f_\theta$ and the pooled visual embeddings to another neural network $g_\phi$. Note that the same networks are used for brand and influencers to learn a common latent space. This parameter sharing is highlighted in the model's architecture depicted in Figure \ref{fig:mt}. In principle, $f_\theta$ and $g_\phi$ have no predefined functional form and all kind of models can be used. In this work, we employed two fully connected neural networks with two hidden layers each. More details on the architecture used in our experiments are provided in Section \ref{sec:exp}.

The new embeddings for each influencer and brand are obtained by multiplying the visual and text representation obtained through $f_\theta$ and $g_\phi$, which are constrained to have the same output dimension, $d_r$:
\begin{align}
    \mathbf{e}_{b} &= f_\theta(\mathbf{b}) \odot g_\phi(\mathbf{b}) \\
    \mathbf{e}_{i,k} &= f_\theta(\mathbf{i}_k) \odot g_\phi(\mathbf{i}_k)  
\end{align}
where $\odot$ is the Hadamard product.
These new embeddings are then fed through a bilinear similarity layer, with a learnable parameter matrix $\mathbf{W}_r \in \real^{d_r\times d_r}$, and a score $z(k=1,2,\dots)$ is computed for each brand and micro-influencer pair as follows: 
\begin{equation}
    z_k = (1 - w_e)(\mathbf{e}_{b}^T\mathbf{W}_r\mathbf{e}_{i,k}) + w_ee
    \label{eq:z_mt}
\end{equation}
where $w_e$ is a trainable weight.
Similarly to what has been described in the previous section, a ranking loss is used and will be referred as $\mathcal{L}_{main}$ hereafter.
Along with the standard ranking task, an auxiliary task has been defined in order to allow the network to learn even better representations. In particular, we introduce a classification loss $\mathcal{L}_{ce}$ (which corresponds to the standard cross-entropy loss) to predict, given $\mathbf{e}_b$ or $\mathbf{e}_{i,k}$ to which macro-category a brand or influencer belong through a shallow neural network $h_\psi$. The overall training loss can be formalized as:
\begin{equation}
    \mathcal{L}(\cdot) = \mathcal{L}_{main}(\cdot) + \lambda\mathcal{L}_{ce,brand}(\cdot) + \gamma\mathcal{L}_{ce, infl}(\cdot)
    \label{eq:mt_loss}
\end{equation}
with $\lambda$ and $\gamma$ two hyperparameters that controls the intensity of the auxiliary tasks for the brand and the micro-influencer being evaluated.
The macro-categories are detailed in Section \ref{sec:dataset} and are one of the key feature available in the dataset used for the experiments.
In the remaining of this work we'll refer to this model as \textbf{WSim-MT} due to the obvious similarities between it and the model described in the previous section. Indeed, they both heavily rely on a learnable bilinear similarity within the model and the final score for the given micro-influencer and brand is computed in almost the same way (see Equation \ref{eq:z} and \ref{eq:z_mt}). 

\subsection{Ranking Loss}
\label{sec:training}
Learning to rank is a well known problem and several loss functions that address it have been defined over the years \cite{liu2011learning}. In this work, given a brand and a pool of $K$ micro-influencers we compute the top one probability of each candidate as in \cite{listnet}: 
\begin{equation}
    \mathbf{p} = \text{softmax}(\mathbf{z}); \qquad \mathbf{z} \in \real^K
    \label{eq:p}
\end{equation}
where $\mathbf{z}$ is a vector whose elements are obtained according to Equation \ref{eq:z} or \ref{eq:z_mt} depending on the model being used. Then the model is trained with the cross entropy loss:
\begin{equation}
    \mathcal{L}_{main}(\mathbf{y},\mathbf{p}) = - \sum\limits_i^{P} y_i log(p_i)
\end{equation}
where $p_i$ is the i-th element of vector $\mathbf{p}$ obtained as per Equation \ref{eq:p} and $y_i=\frac{1}{|\mathcal{I}_+(\mathbf{b})|}$ if micro-influencer $i$ is associated with brand $\mathbf{b}$ or 0 otherwise.
In order to exploit the rank information of partial sequences we employed a simplified variant of the approach describe in \cite{MIR}. In particular, given a brand $\mathbf{b}$, $P$ pools of $K$ candidate micro-influencers are created, such that in each pool a varying number of positive micro-influencers is always present. In particular, the number of positive influencer in a pool can range between 1 and $K$. Negative examples within the pool are randomly sampled without replacement from the dataset.
In order to use all the valuable information provided within the dataset, the number of pools $P$ scales with the amount of positive examples for each given brand. For example, if for a given brand there are 5 positive examples, and the pool size is $k=3$, then 15 pools will be created: 5 pools containing only one positive instance, 5 pools containing 2 positive instances and 5 pools with all positive instances. In this way, we guarantee that all the positive examples are used within the training procedure while keeping the computation tractable. This would not happen if $P$ would contain all the possible permutations of the partial sequences of positive examples, as the number of pools for a given brand $\mathbf{b}$ would be: $\sum\limits_{k=1}^K \binom{|\mathcal{I}_+(\mathbf{b})|}{k}$ which can quickly become intractable.
When employing the "partial-sequence mode" instead of the full "listwise" one, the definition of $y_i$ change slightly. The normalizing factor is not the number of positive examples associated with a brand anymore, instead it is the number of positive examples associated with a brand within a given pool and it can range between 1 and $K$.

\section{Experiments}
\label{sec:full_exp}
\subsection{Dataset}
\label{sec:dataset}
Open Source datasets for micro-influencer recommendation are hard to find and expensive to create, so there are a limited set of baseline tasks to choose from. Recently, the authors of \cite{MIR} shared the first brand, micro-influencer multimedia dataset to spark research in the field. We therefore evaluate our methods on it in this work.

The dataset targets brand accounts on Instagram belonging to one of 12 macro categories: Airline, Auto, Clothing, Drink, Electronics, Entertainment, Food, Jewellery, Makeup, Non-Profit, Shoes and Services. 30 brands are selected for each category for a total of 360 brands. A micro-influencer is assumed to be associated with a brand when it is cited in one of the last 1000 posts of the brand. Moreover, to be eligible, a micro-influencer should have a number of followers ranging between 5000 and 100000. On average, each brand is associated with 11 micro-influencers. 
For each brand and micro-influencer in the dataset a mix of profile and posts-related information is available. In particular, the images, text, number of likes/comments of their last 50 posts is available ($N_p=50$). 

\subsection{Experimental Setup}
\label{sec:exp}
We performed 5-fold cross validation to remain consistent with the 80-20 split performed in \cite{MIR} and ensure a fair comparison. We enforce that all brand categories are available in each train-test split.
The evaluation is done by scoring a brand against all the candidate micro-influencers. The obtained scores are then sorted in descending order to obtain the brand-specific micro-influencer ranking.

We trained all our models in an end-to-end fashion using the Adam optimiser \cite{adam} for 200 epochs with Early Stopping (where 20\% of the samples in the training set were held out and used for validation purposes) and a learning rate of 0.001.
The visual and textual embeddings' size obtained from the pretrained models are 25088 and 300 respectively. In \textbf{WSim-MT} $f_\theta$ and $g_\phi$ are fully connected neural networks with two hidden layers composed by (300, 512) and (4096, 512) hidden units respectively.
We used ReLU activation and a droupout rate of 0.5, the intensity regularizer $\lambda$ and $\gamma$ described in Equation \ref{eq:mt_loss} for the multi-task model are both set to 0.5.
Also, $h_\psi$ is an affine transformation followed by a softmax activation.

\subsection{Results}
\begin{table*}[]
\centering
\caption{Comparison of our methods with the baselines in terms of performance and model complexity. The results comes from a 5 fold cross validation except for one case where the results has been reported as described in \cite{MIR}. The number of parameters in the Table does not include the ones of the pretrained models (VGG-16 and the spaCy language model).}
\begin{tabular}{lccccc}
\toprule
             & \textbf{AUC}        & \textbf{Recall@10}                          & \textbf{Recall@50}         & \textbf{MedR}  & \textbf{N.Params}           \\
\midrule
Random.      & 0.493 $\pm$ 0.003   & 0.007 $\pm$ 0.002                  & 0.041 $\pm$ 0.004 & 71.4 $\pm$ 5.3 & 0\\
SimCos & 0.755 $\pm$ 0.006   & 0.156 $\pm$ 0.007                  & 0.368 $\pm$ 0.009 & 2.5 $\pm$ 0.5  & 0\\
\midrule
MIR(k=4)     & 0.821 $\pm$ 0.008   & 0.082 $\pm$ 0.006                 & 0.307 $\pm$ 0.014 & 8.7 $\pm$ 1.3  & $\sim$105M\\
MIR(from \cite{MIR})     & 0.849   & 0.135                              & 0.428             & 6             & -\\
\midrule
\textbf{WSim(k=4) }   & 0.858 $\pm$ 0.007   & 0.175 $\pm$ 0.012                  & 0.467 $\pm$ 0.020 & 3.0 $\pm$ 0.6  & $\sim$\textbf{25K}\\
\textbf{WSim(k=6)}    & 0.861 $\pm$ 0.008   & 0.178 $\pm$ 0.013                  & 0.469 $\pm$ 0.023 & 2.9 $\pm$ 0.5  & $\sim$\textbf{25K}\\
\textbf{WSim-MT(k=4)} & 0.890 $\pm$ 0.009   & 0.163 $\pm$ 0.004                  & 0.479 $\pm$ 0.018 & 2.2 $\pm$ 0.4  & $\sim$105M\\
\textbf{WSim-MT(k=6)} & \textbf{0.920 $\pm$ 0.014}   & \textbf{0.193 $\pm$ 0.011}                  & \textbf{0.548 $\pm$ 0.037} & \textbf{2.0 $\pm$ 0.4}  & $\sim$105M\\
\bottomrule
\end{tabular}
\label{tab:results}
\end{table*}

In this section we will compare the methods introduced in Section \ref{sec:wsim} with several baselines, some of which represent the state-of-the-art in brand/micro-influencers recommendations. The models considered for comparison are:
\label{subsec:baselines}
\begin{itemize}
    \item \textbf{Random}: given a brand, the associated list of micro-influencers are randomly sampled among the available ones.
    \item \textbf{SimCos}: concatenate the images and text embeddings to obtain a unique vector representation for brand and micro-influencers. Once the concatenated vectors are obtained, each micro-influencer is scored according to its cosine similarity w.r.t the brand representation. 
    \item \textbf{MIR(k)}: The method proposed in \cite{MIR}. Differently from our methods, no trainable similarity is used between micro-influencer and brand; moreover, the engagement score is combined with the brand/micro-influencer similarity in a static, non trainable manner. Finally, multi-task learning was not considered in this previous work.
\end{itemize}

To remain consistent with previous work we employed four main metrics to evaluate our models. \textit{AUC}, that measures the probability of a positive example being ranked higher then a negative one. \textit{Recall @ k}, with $k \in \{10,50\}$ as a classic recommender system metric; it considers the fraction of positive examples, among all the available one for a given brand, that are included in the top-k ranked items. Finally, \textit{MedR} considers the median position of the highest ranked positive example for a given brand in the test set.

Table \ref{tab:results} summarizes the results obtained by our methods compared against the previous ones.
For \textbf{MIR(k)}, we report both the results presented in the original paper (the ones with no uncertainty) and the results obtained via our implementation. Some discrepancy is expected as our implementation has been tested on multiple different test datasets.

We notice that the cosine similarity model, despite it's simplicity, already provides a strong baseline. It is therefore reasonable to expect \textbf{WSim}, a straightforward generalization of the previous model, to perform better.
Indeed, by introducing minimal complexity into the model, we can move from a task-agnostic model to one which can specialize for the task at hand by optimizing a principled objective.  This improvement is clear in Table \ref{tab:results}, where we can also observe some kind of insensitivity to the size of the pool of candidates micro-influencers used during training.

\begin{figure}
    \centering
    \includegraphics[width=\linewidth]{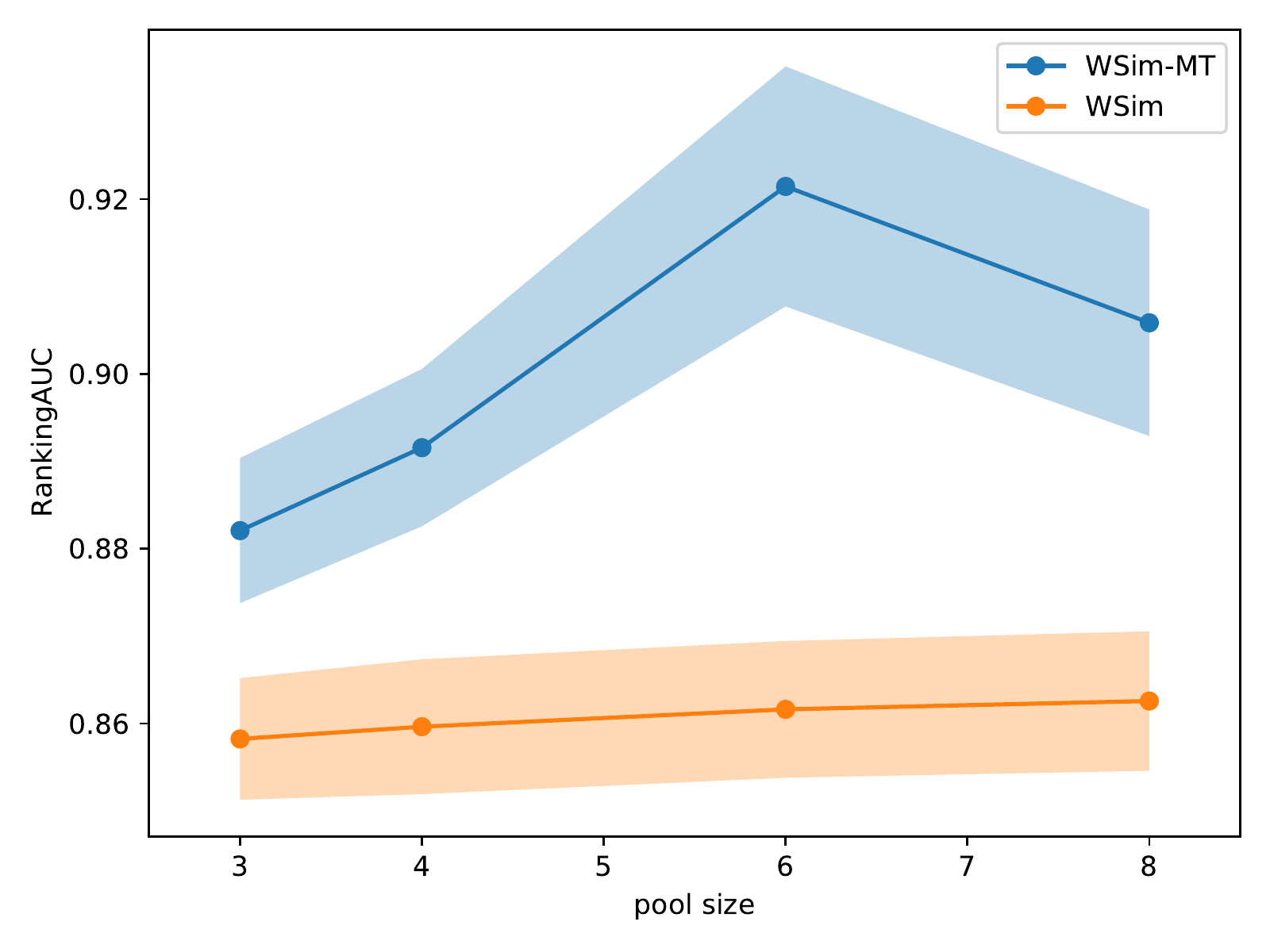}
    \caption{RankingAUC as a function of the pol size for both WSim and Wsim-MT.}
    \label{fig:auc_k}
\end{figure}

\textbf{WSim-MT} increases the complexity of the model, but gives a large improvement in all the considered metrics, in particular we improved by 8.2\% in AUC, 42.9\% in Recall @ 10 and 28\% in Recall @50 compared to the previous state-of-the-art. We note that in \textbf{WSim} the use of a larger pool size has no significant impact on performance. It can be seen in Figure \ref{fig:auc_k} that the algorithm is almost insensitive to this kind of change. The same does not hold for \textbf{WSim-MT}, where using a pool size of $k=6$ provide the best results overall. We also note that continuously increasing the pool size at some point has a negative impact on performance. We attribute this behaviour to the sampling mechanism we adopted. Indeed, in order to keep the computation tractable, the number of samples that we use for training scales linearly with the pool size. This is not enough for larger pool sizes, as the model would likely need many more samples to learn an optimal ranking in this case.  

\section{Interpretation}
\label{sec:interpretation}

Up to this point, we have introduced influencer ranking algorithms that perform well on overall metrics. However, as these techniques rely heavily on deep neural networks, it can be difficult for a human to understand why the posts of a particular influencer lead to its ranking position for a certain brand. Being able to better interpret why an influencer is matched well with a brand can be of interest to both brands and influencers. Along with providing verification that the ranking algorithm is functioning as expected, this information can help influencers or brands tailor their content to better match with the content of a social media account that has an audience that they wish to target.

As our dataset is based on Instagram posts, in which images are predominant, we focus on interpreting the visual content for individual posts that could be of interest to a particular brand or influencer. To understand which components of an image are important, we introduce a technique inspired by the Grad-CAM algorithm~\cite{selvaraju2017grad}, which operates on image classification networks, highlighting the parts of the input image that are most relevant to the output classification. Grad-CAM is designed to work on any convolutional neural network by determining an importance score across all the dimensions in the final convolutional layer of the network. These importances can be turned into a heatmap that can be superimposed over the input image by summing over the non-spatial dimensions of the feature space. This heatmap then highlights the most relevant areas for the final classification. In the original algorithm the importance scores are determined by the size of the gradients of the top predicted class with respect to the activations in the last convolutional layer. 

\begin{figure}
     \centering
     \begin{subfigure}
         \centering
         \includegraphics[width=0.2253\textwidth]{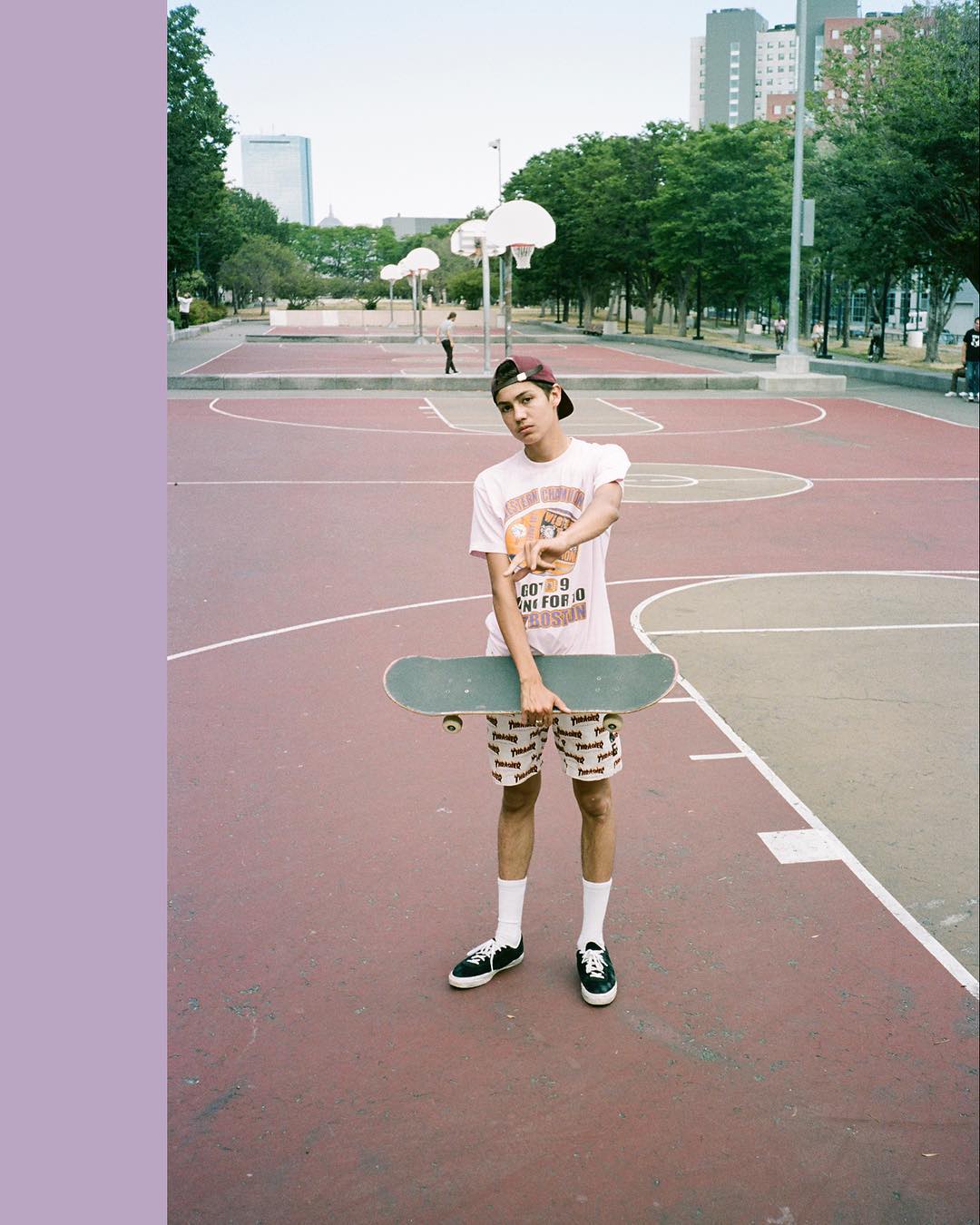}
         \label{fig:nike1}
     \end{subfigure}
     \begin{subfigure}
         \centering
         \includegraphics[width=0.2253\textwidth]{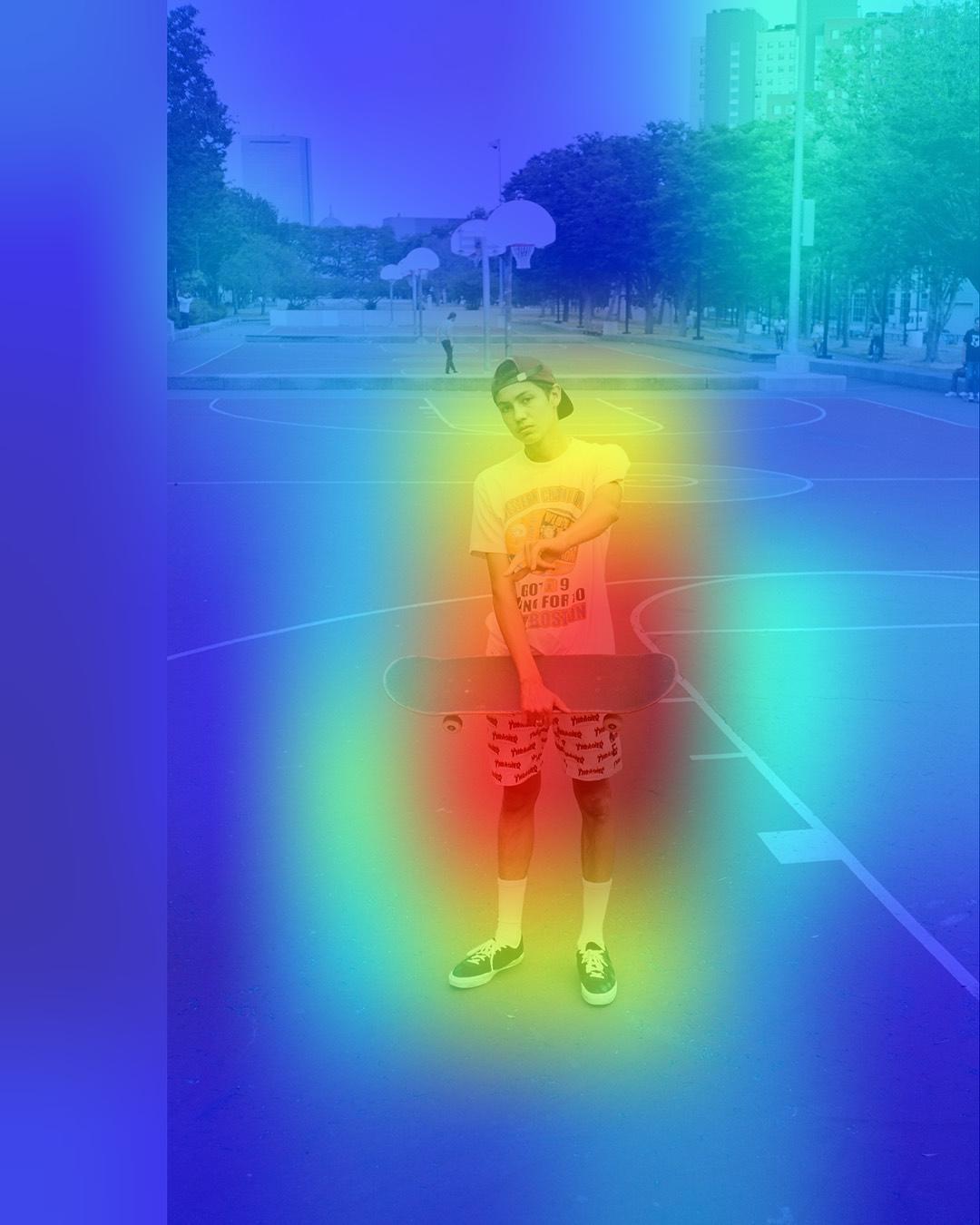}
         \label{fig:trained_plus_vgg_food3}
     \end{subfigure}
     \\
     \begin{subfigure}
         \centering
         \includegraphics[width=0.2253\textwidth]{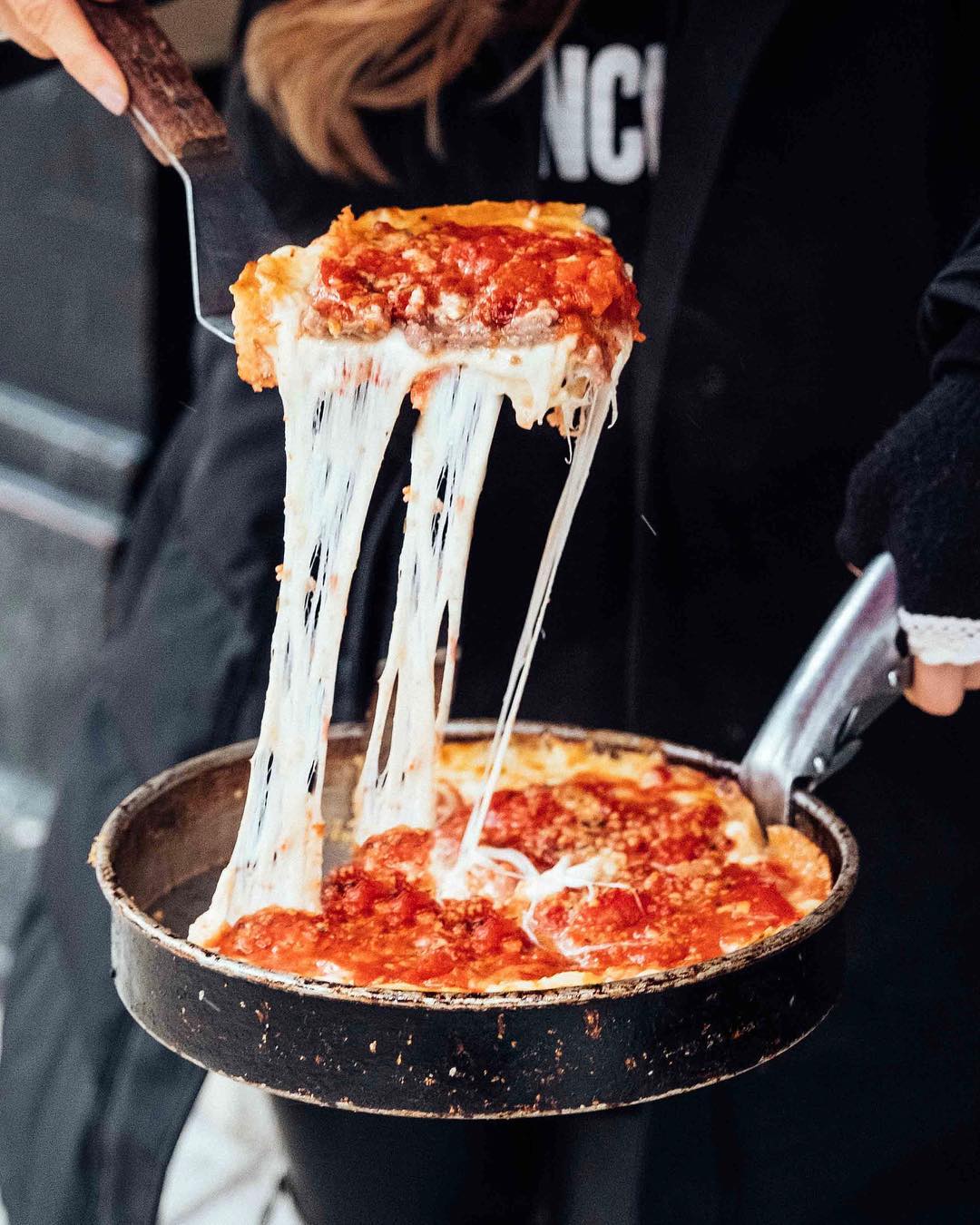}
         \label{fig:trained_plus_vgg_nike1}
     \end{subfigure}
     \begin{subfigure}
         \centering
         \includegraphics[width=0.2253\textwidth]{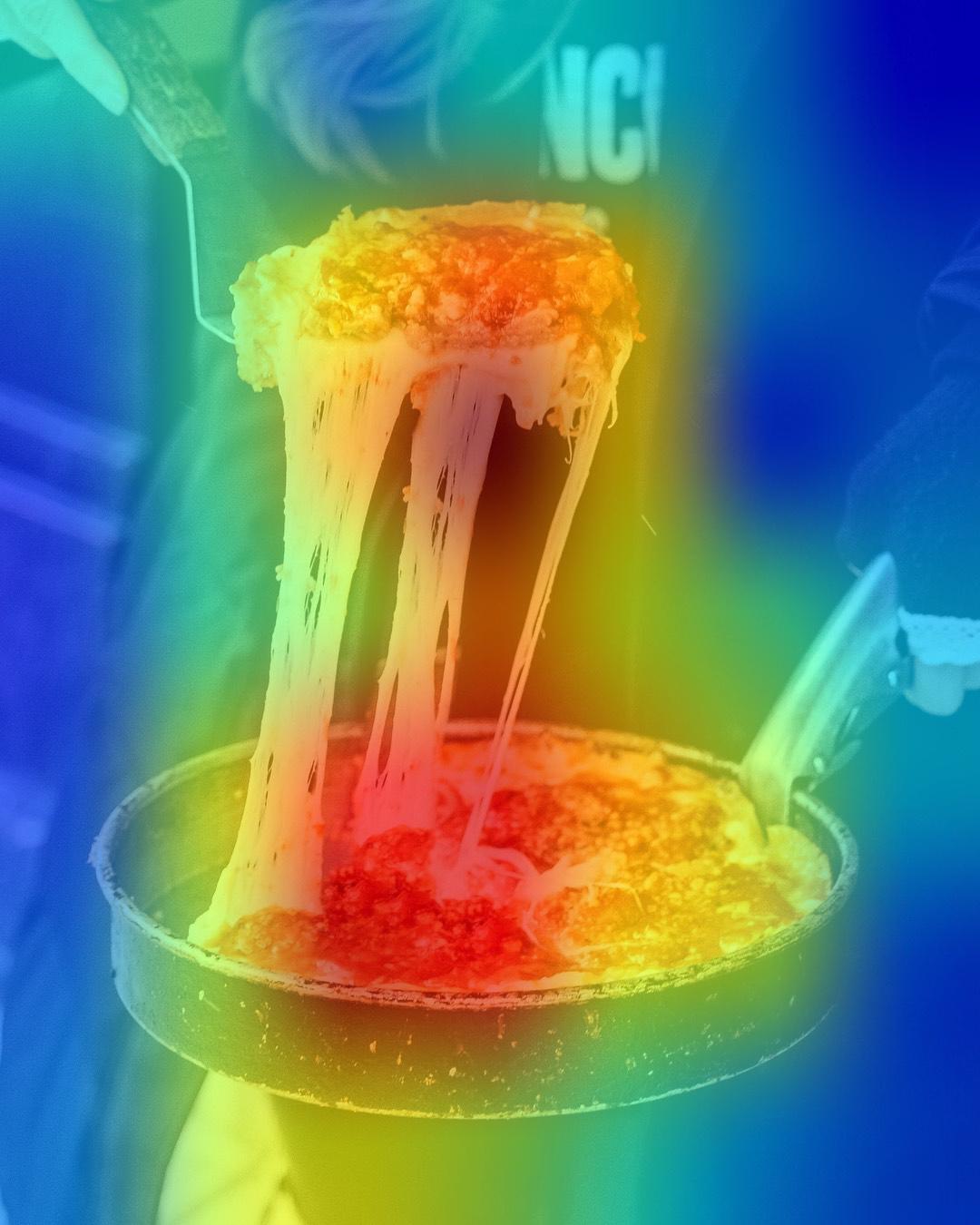}
         \label{fig:trained_plus_vgg_food3}
     \end{subfigure}
        \caption{Heatmaps of general image features the influencer ranking network (WSim) gives most importance to when comparing image similarities. Shown for a post from a fashion influencer (top) and a food influencer (bottom), where red is more important}
        \label{fig:trained_plus_vgg}
\end{figure}

We adapt these ideas to our use case by introducing a new measure of importance based on the architecture of the \textbf{WSim} model introduced in Section~\ref{sec:wsim}. In this model, input images are passed through VGG-16 and activations of the final convolutional layer are used as a feature representation, $\mathbf{X}_{\text{vgg}} \in \real^{s_1\times s_2 \times f_{N}}$, where $s_1$ and $s_2$ are the two spatial dimensions of the image and $f_{N}$ is the dimension of convolutional filter channels in the final layer. As described earlier, the visual image dimension, $\dv$ is the unrolled values of $\mathbf{X}_{\text{vgg}}$, such that:
\begin{equation}
    \dv \equiv s_1\times s_2 \times f_{N}
\end{equation}
These representations are pooled across posts for a brand or influencer and a similarity score between the two obtained through a trained bilinear similarity, which is parameterised by a diagonal matrix $\mathbf{W}_v \in \real^{\dv\times\dv}$. Within this matrix there are $\dv$ parameters, which are one-to-one with the parameters in $\mathbf{X}_{\text{vgg}}$, with a magnitude that determines how relevant particular features are when determining the similarity of two images. They therefore provide a natural measure of importance,  $\mathbf{I}_v \in \real^{\dv}$, for each component of the visual features $\mathbf{X}_{\text{vgg}}$:
\begin{equation}
    \mathbf{I}_v \equiv \text{diag}(\mathbf{W}_v)
\end{equation}
We can now use $\mathbf{I}_v$ to replace the gradients used in Grad-CAM, while allowing a heatmap for a particular image, $\mathbf{H}\in\real^{s_1 \times s_2}$, to be built in the same way:
\begin{equation}
    \label{eq:heatmap}
    \mathbf{H} = \sum\limits_{i}^{f_N} \mathbf{I}_v \odot \mathbf{X}_{\text{vgg}}
\end{equation}
where the unrolled components of $\mathbf{X}_{\text{vgg}}$ are one-to-one with the components of $\mathbf{I}_v$ and the non-spatial dimensions are summed over.
This heatmap can be used to visualise the components of an image that are most relevant to the \textbf{WSim} model when calculating the similarity of any two accounts. This process applied to the images from two different posts from fashion and food influencers can be seen in Fig.~\ref{fig:trained_plus_vgg}. In the food focused post,  we can see that the model focuses on the pizza and ignores the background. Other posts would be scored as being similar to this post if they had similarly predominant food based visual components. The same can be said for the person and their clothes in the fashion focused post. This helps to give us confidence that our model is performing as expected, paying most attention to the parts of the post a human would have deemed most important. 

\begin{figure}
     \centering
     \begin{subfigure}
         \centering
         \includegraphics[width=0.23\textwidth]{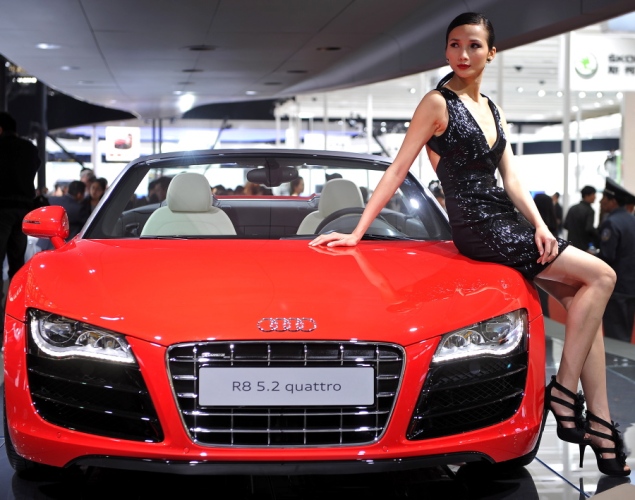}
         \label{fig:orig_car_model2}
     \end{subfigure}
     \\
     \begin{subfigure}
         \centering
         \includegraphics[width=0.23\textwidth]{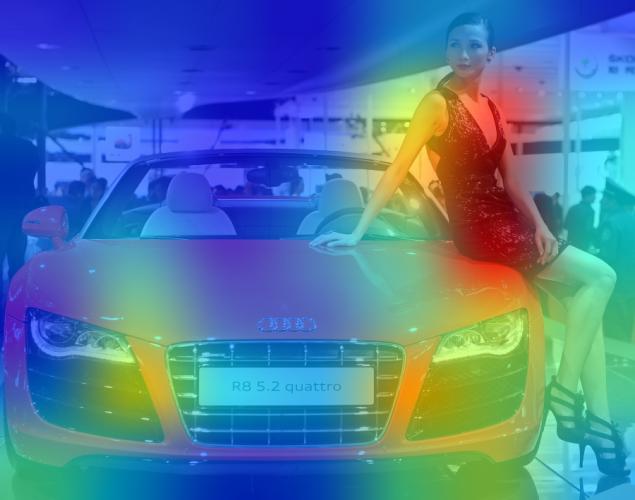}
         \label{fig:ootdfash_car_model2}
     \end{subfigure}
     \hfill
     \begin{subfigure}
         \centering
         \includegraphics[width=0.23\textwidth]{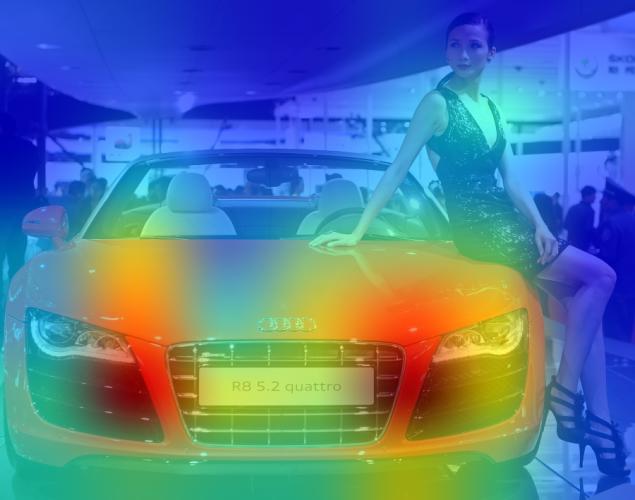}
         \label{fig:audi_car_model2}
     \end{subfigure}
        \caption{Heatmaps of brand specific image features the influencer ranking network (WSim) gives most importance to. On top is the original image, below is the importance for a fashion brand (bottom left) and a car brand (bottom right), where red is more important.}
        \label{fig:car_model2}
\end{figure}

\begin{figure}
     \centering
     \begin{subfigure}
         \centering
         \includegraphics[width=0.21\textwidth]{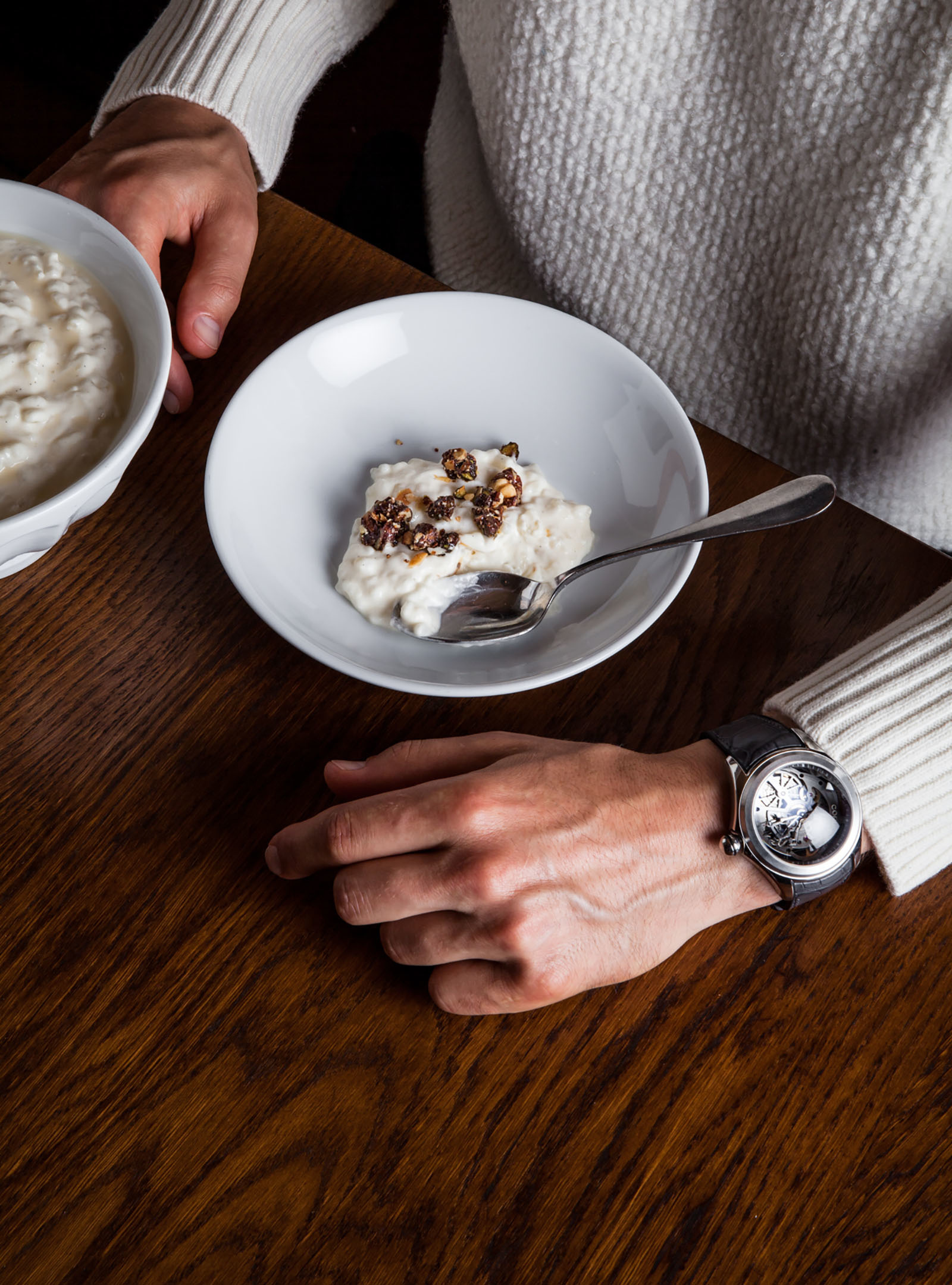}
         \label{fig:orig_watch_food}
     \end{subfigure}
     \\
     \begin{subfigure}
         \centering
         \includegraphics[width=0.21\textwidth]{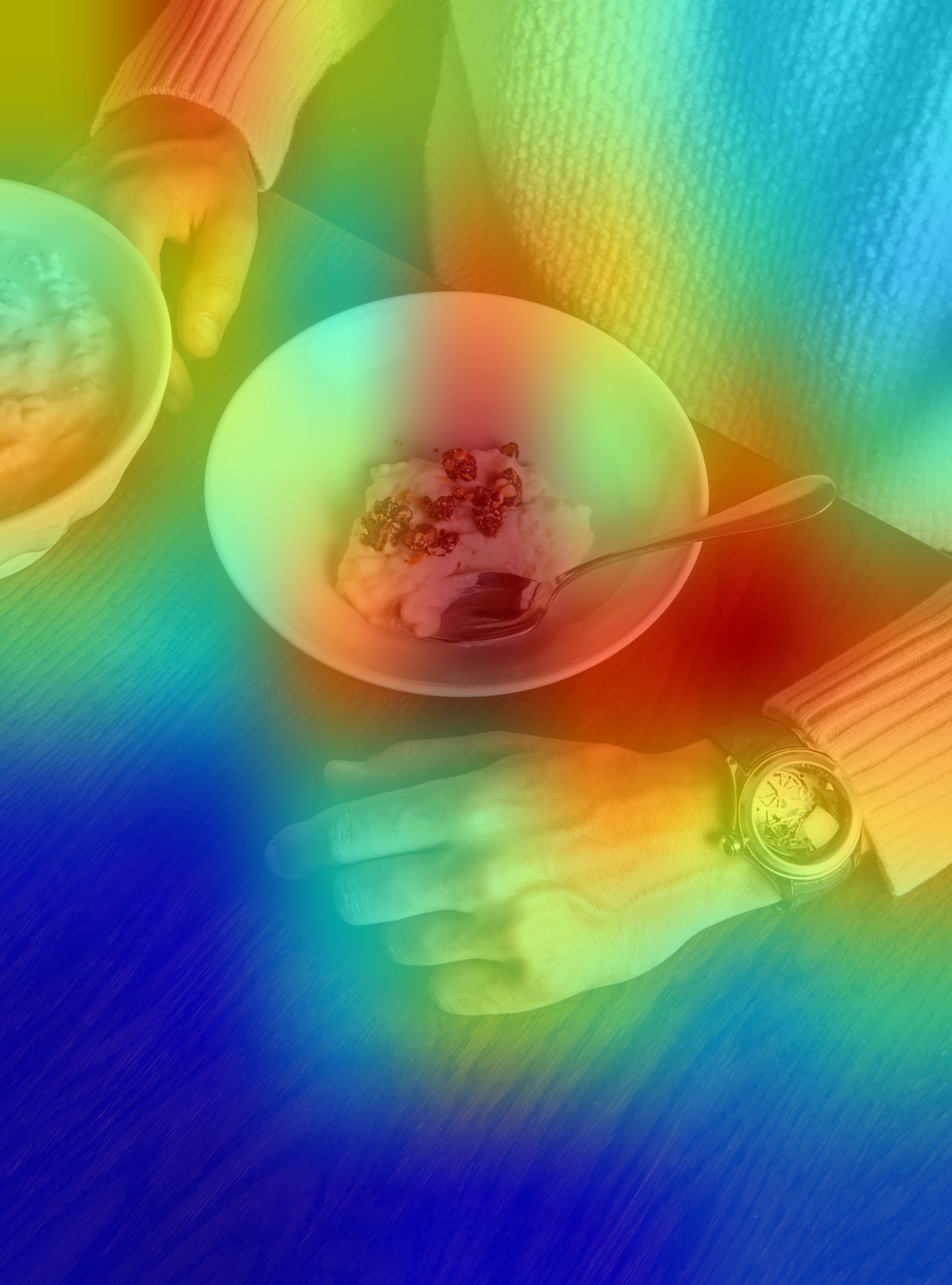}
         \label{fig:shake_shack_watch_food}
     \end{subfigure}
     \begin{subfigure}
         \centering
         \includegraphics[width=0.21\textwidth]{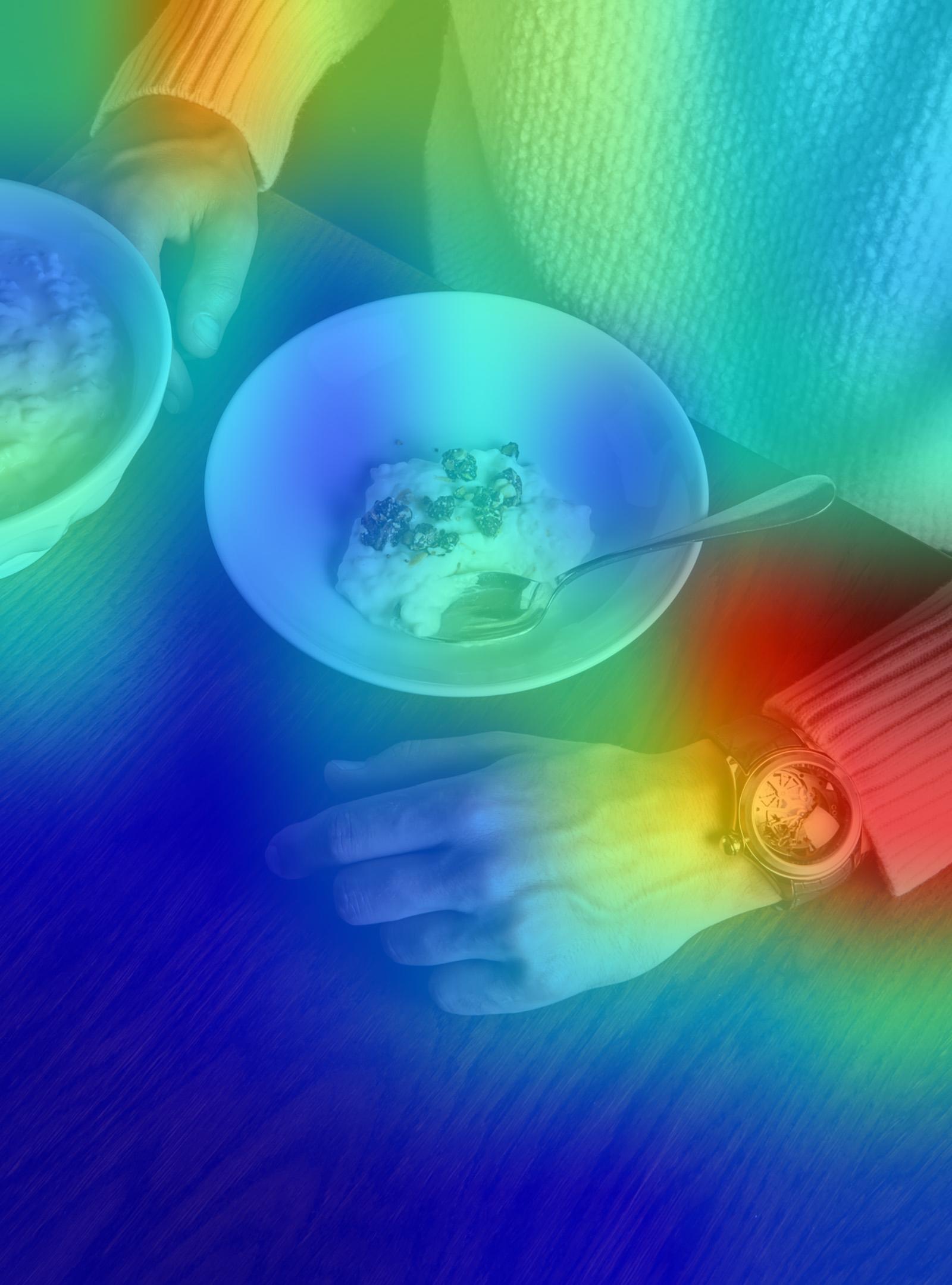}
         \label{fig:briston_watches_watch_food}
     \end{subfigure}
        \caption{Heatmaps of brand specific image features the influencer ranking network (WSim) gives most importance to. On top is the original, below is the importance for a food brand (bottom left) and a jewellery brand (bottom right), where red is more important.}
        \label{fig:watch_food}
\end{figure}

The importance score $\mathbf{I}_v$ just highlights the general purpose features that are important for calculating the similarity of images. It can be made more specific to a particular brand or influencer by multiplying the bilinear similarity matrix by the pooled feature vector of the brand, $\mathbf{b}_v$, or influencer, $\mathbf{i}_v$, to get a new importance vector, $\mathbf{I}^b_v\in \real^{\dv}$ in the case of a brand: 
\begin{align*}
    \mathbf{I}^b_v &\equiv \mathbf{W}_v\mathbf{b}_v
\end{align*}
This importance score is again one-to-one with the final convolutional layer of VGG-16 and can be used to derive a heatmap that highlights the components of an image most relevant to the brand or influencer in question, directly replacing $\mathbf{I}_v$ with $\mathbf{I}^b_v$ in Equation~\ref{eq:heatmap}. This can be seen in Fig.~\ref{fig:car_model2}, where the importance heatmap of an image containing a woman and a car has been calculated with $\mathbf{b}_v$ taken from either a fashion brand, or a car brand. In the case of the fashion brand, this importance measure suggests that the model pays most attention to the woman when calculating similarity. For the car brand the model pays most attention to the car. A similar comparison is carried out in Fig.~\ref{fig:watch_food} for a food and jewellery brand. In all these cases it is clear that the model is paying attention to the part of the image one would naively have thought most relevant to the brand in question. 

These techniques for interpreting the $\textbf{WSim}$ model help us to confirm the model is behaving as intuitively expected. However, they can also be used by brands or influencers that wish to analyse their media content. When deciding on images for new posts, for example, these techniques can help to highlight which components of new images are most similar to previous posts made by the brand. This allows a brand media manager to understand and direct their visual content. Such techniques could also be used to help brands reach new audiences, by choosing images most similar to the posts made by an influencer with an audience a brand would like to target.

\section{Conclusions and future work}
\label{sec:conclusions}

We have improved on the state-of-the-art in micro-influencer ranking, building on the work originally presented in \cite{MIR}. This was achieved by introducing two novel deep neural network architectures based on a trainable inner product (\textbf{WSim}), which drastically reduces the number of trainable parameters required, or the introduction of multi-task learning (\textbf{WSim-MT}), which performs best overall. 

On top of this, inspired by Grad-CAM \cite{selvaraju2017grad}, we make use of the architecture of the \textbf{WSim} model to introduce a layer of interpretability of the ranking decisions made by our models. This has helped us validate that the ranking is taking into account the features we would expect and provides tools for brands and influencers to improve their media content in the future.

It is worth noting that this work focuses on micro-influencers, but the tools developed, particularly the interpretability layer, can also be useful for influencers with large followings. These techniques can also be generalised to arbitrary multimedia ranking tasks that have datasets with a similar structure.

To expand on this work further, it would be interesting to test the performance on other social media platforms in which text is more relevant. If it works well, the techniques for interpretability could be expanded to text, allowing brands to isolate key words and phrases that help target particular audiences. Additionally, it could be interesting to investigate if the neural network architectures presented could be used for predicting the engagement of individual posts. In the case that this is successful, the interpretability layer could help brands isolate the components of images that are most important for producing engaging content.



\bibliography{sample-authordraft}
\bibliographystyle{ieeetr}




\end{document}